\documentclass[letterpaper, 10 pt, journal, twoside]{IEEEtran}

\usepackage{graphics} 
\usepackage{epsfig} 
\usepackage{mathptmx} 
\usepackage{times} 
\usepackage{amsmath} 
\usepackage{amssymb}  
\usepackage{booktabs} 
\usepackage{cite}
\usepackage{amsfonts}
\usepackage{amsbsy}
\usepackage{placeins}
\usepackage{multirow}
\usepackage{tabularx}
\usepackage{bm}

\DeclareMathOperator*{\argmin}{arg\,min}

\begin{document}

\title{Control Pneumatic Soft Bending Actuator with Feedforward Hysteresis Compensation by Pneumatic Physical Reservoir Computing}

\author{Junyi Shen, Tetsuro Miyazaki, \textit{Member, IEEE}, and Kenji Kawashima, \textit{Member, IEEE}
\thanks{Preprint version of the IEEE Robotics and Automation Letters 10.1109/LRA.2024.3523229, final version can be found on the IEEE Xplore.}\thanks{This work was supported by JSPS KAKENHI 21H04544 and Bridgestone Corporation. \textit{(Corresponding authors: Kenji Kawashima, Junyi Shen)}}\thanks{Junyi Shen, Tetsuro Miyazaki, and Kenji Kawashima are with Department of Information Physics and Computing, the University of Tokyo, 113-8656, Tokyo, Japan (e-mail: junyi-shen@g.ecc.u-tokyo.ac.jp; tetsuro\_miyazaki@ipc.i.u-tokyo.ac.jp; kenji\_kawashima@ipc.i.u-tokyo.ac.jp}\thanks{Digital Object Identifier (DOI): see top of this page.}
}

\markboth{IEEE Robotics and Automation Letters. Preprint Version. Accepted December, 2024}
{Shen \MakeLowercase{\textit{et al.}}: Control Pneumatic Soft Actuator by Pneumatic Physical Reservoir Computing} 

\maketitle

\begin{abstract}
The nonlinearities of soft robots bring control challenges like hysteresis but also provide them with computational capacities. This paper introduces a fuzzy pneumatic physical reservoir computing (FPRC) model for feedforward hysteresis compensation in motion tracking control of soft actuators. Our method utilizes a pneumatic bending actuator as a physical reservoir with nonlinear computing capacities to control another pneumatic bending actuator. The FPRC model employs a Takagi-Sugeno (T-S) fuzzy logic to process outputs from the physical reservoir. The proposed FPRC model shows equivalent training performance to an Echo State Network (ESN) model, whereas it exhibits better test accuracies with significantly reduced execution time. Experiments validate the FPRC model's effectiveness in controlling the bending motion of a pneumatic soft actuator with open-loop and closed-loop control system setups. The proposed FPRC model's robustness against environmental disturbances has also been experimentally verified. To the authors' knowledge, this is the first implementation of a physical system in the feedforward hysteresis compensation model for controlling soft actuators. This study is expected to advance physical reservoir computing in nonlinear control applications and extend the feedforward hysteresis compensation methods for controlling soft actuators.
\end{abstract}


\section{Introduction}

\IEEEPARstart{P}{neumatic} actuation is a widely favored driving method in soft robotics. Pneumatic Artificial Muscles (PAMs), with the renowned McKibben design \cite{chou1996measurement}, can produce large force with low self-weights. While PAMs are designed to execute linear motions, modifications like incorporating physical constraints into the standard PAM structure \cite{shen2023trajectory} have enabled PAM-based soft actuators to perform bending movements.

Their flexible materials offer soft robots benefits like user-friendliness but also bring them significant control challenges \cite{nakajima2015information}, particularly when precision and rapid response are required. One of the problems is hysteresis, which is a time-dependent nonlinearity caused by the Coulomb friction between mechanical components and the viscoelasticity of the soft materials \cite{chou1996measurement}. Hysteresis causes nonlinear loops with dead zones in soft actuators' input-to-output relationships \cite{shen2023trajectory}, leading to difficulties in precisely controlling their motions. 

As a result, one popular method to control soft actuators is integrating a feedforward hysteresis compensation model into a feedback control loop \cite{xie2018hysteresis, jiang2022intelligent}. The feedforward hysteresis compensation models are often built by the combination of a group of mathematical play-operators \cite{xie2018hysteresis} or using Recurrent Neural Networks (RNNs) \cite{jiang2022intelligent}. RNN-based modelings can capture complex time-dependent dynamics and feature easy implementation. However, neural networks typically have poor interpretability and require a large amount of data for training. Meanwhile, their gradient-based training approaches can be computationally expensive. Compared with RNNs, hysteresis models based on mathematical play-operators possess better transparency and reasonable physical interpretability \cite{wang2020new}. Nonetheless, their development usually depends on complex and time-consuming optimization techniques, such as genetic algorithms (GA) and particle swarm optimization (PSO) \cite{hassani2014survey}. 

In light of classic RNNs' computational complexity, Reservoir Computing (RC) \cite{jaeger2002adaptive}, as a variant of RNNs, has been introduced to predict complex nonlinear behaviors. In RC models like the widely used Echo State Network (ESN) \cite{jaeger2002adaptive}, the input signal is projected into a high-dimensional reservoir state via an input layer. The state is updated by a large and randomly initialized network named the `reservoir.' Notably, the input layer and the reservoir are fixed during training, with only the output layer, which handles the reservoir state by linear weights, being trained with linear regressions. As it requires no adjustment, the reservoir can be replaced by physical systems like soft bodies \cite{nakajima2015information}, thus giving rise to physical reservoir computing (PRC). Since physical systems' state update does not involve the large `reservoir' used in ESNs—which may contain tens to thousands of nodes \cite{park2016online}—PRC models can show greatly reduced computational requirements in their execution \cite{akashi2024embedding}, making them suitable for real-time control applications.

PRCs can be used with open-loop and closed-loop forms. Prior research has employed the closed-loop PRC to generate autonomous behaviors of soft robots (or actuators) \cite{akashi2024embedding, eder2018morphological} and the open-loop PRC for sensory feedback in control systems \cite{hayashi2022}, as well as other real-time computational tasks \cite{nakajima2015information}. However, the application of open-loop PRC in feedforward hysteresis compensation models for controlling soft actuators remains unexplored. Furthermore, the PRC, which relies on physical systems with actual existences, may be susceptible to real-world disturbances. While the robustness of the closed-loop PRC under such conditions has been examined in \cite{eder2018morphological}, the impact of environmental disturbances on the performance of open-loop PRC has not been investigated in \cite{nakajima2015information} and \cite{hayashi2022}.

The Takagi-Sugeno (T-S) fuzzy logic \cite{takagi1985fuzzy} is a powerful tool in nonlinear system identifications \cite{cheng2016adaptive}. It has been utilized for precise hysteresis modeling with the nonlinear autoregressive moving average (NARMA) model \cite{cheng2016adaptive}. The T-S fuzzy logic has also been integrated into a multi-ESN model for enhancing the performance in identifying nonlinear systems \cite{mahmoud2022tsk}. Hence, we consider the T-S fuzzy logic also capable of improving the PRC's capabilities in modeling complex nonlinear dynamics.

This paper proposes a novel control approach using PRC in a feedforward hysteresis compensation model to control a soft bending actuator. Our method employs a dual-PAM actuator with two air chambers as a pneumatic physical reservoir. A T-S fuzzy model is applied to process the reservoir's readouts and generate the feedforward hysteresis compensation model's output. The proposed method shows better accuracies than an ESN and significantly reduced execution time in the model test phase. Experimental results confirm the proposed model's effectiveness in open and closed-loop bending motion tracking control. The proposed model's strong robustness against real-world disturbances has also been experimentally demonstrated.

The remainder unfolds as follows: Section 2 presents the design and hysteresis of the pneumatic soft actuator; Section 3 describes the feedforward hysteresis compensation models; Section 4 exhibits experimental results; Section 5 concludes.

\section{Pneumatic Bending Actuators}

\subsection{Structural Design}

\begin{figure}[tb]
    \centering
\includegraphics[width=0.85\linewidth]{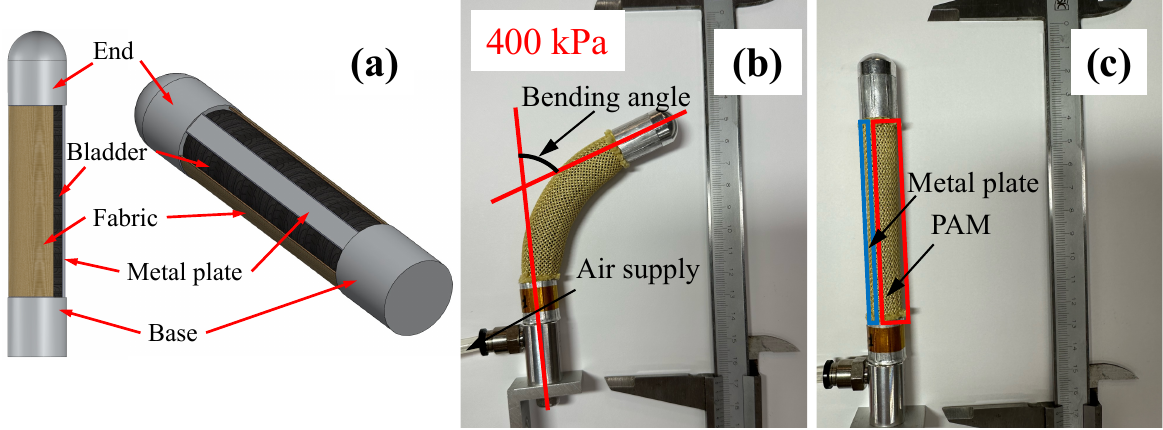}
    \caption{Pneumatic soft bending actuator: (a) Structure; (b) Bending under pressurization; (c) Returning to initial straight shape upon depressurization.}
    \label{fig: pam}
    \vspace{-0.2cm}
\end{figure}

Fig.~\ref{fig: pam} shows the pneumatic soft bending actuator, whose motion is this work's control object. Its structure comprises a metallic base, a rubber bladder, a metal end, a linking metal plate, and an inextensible fabric shield. Upon pressurization, the rubber bladder expands radially and contracts axially, following the mechanism of PAMs. The metal plate restricts the actuator's uniform axial contraction, thus inducing its bending, as Fig.~\ref{fig: pam} (b) illustrates. Upon depressurization, the actuator returns to its straight shape, as depicted in Fig.~\ref{fig: pam} (c).

\begin{figure}[tb]
    \centering
\includegraphics[width=0.85\linewidth]{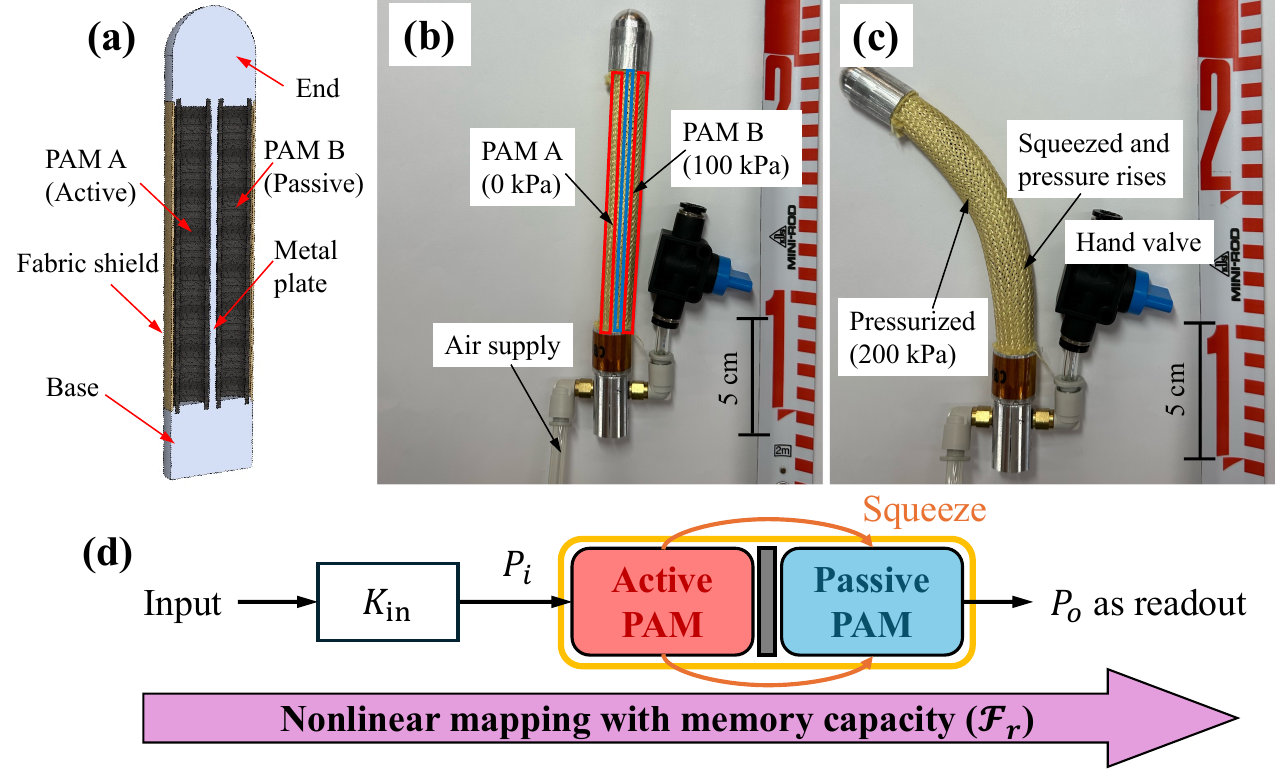}
    \caption{Dual-PAM  actuator as a physical reservoir: (a) Structural design; (b) Atmospheric pressure in active PAM (PAM A) and pre-pressurized state in passive PAM (PAM B); (c) Actuation of PAM A and consequential pressure increase in compressed PAM B. (d) Schematic of the nonlinear mapping.}
    \label{fig: reservoir}
    \vspace{-0.2cm}
\end{figure}

The dual-PAM actuator serving as the pneumatic physical reservoir in this work is shown in Fig.~\ref{fig: reservoir}. Its structure shown in (a) closely mirrors the design in Fig.~\ref{fig: pam} (a), with a second rubber bladder included and a centrally placed metal plate flanked by two antagonistically positioned air chambers. In the physical reservoir setup, one chamber (PAM A in Fig.~\ref{fig: reservoir}) functions as an active PAM actuated by the designed pressure input, the other chamber (PAM B in Fig.~\ref{fig: reservoir}) acts as a passive PAM initially pressurized to 100 kPa and closely sealed by a hand valve, as shown in Fig.~\ref{fig: reservoir} (b). In controlling the bending motion of the soft actuator shown in Fig.~\ref{fig: pam}, the active PAM is pressurized by an actuation input $P_i$ [kPa] expressed by:
\begin{equation}
    P_i (k) = K_\text{in} \theta_d (k),
\label{eq: convert}
\end{equation}
where $\theta_d$ [deg] is the actuator's desired bending angle and $K_\text{in}$ [kPa/deg] is a convert factor. The physical system needs to have nonlinearity, memory capacity, and high dimensionality for functioning as a physical reservoir \cite{nakajima2015information}. In the proposed setup, actuation and inflation of the active PAM cause the inextensible fabric shield squeezing the pre-pressurized and sealed passive PAM, thereby altering its inner pressure $P_o$ [kPa], as shown in Fig.~\ref{fig: reservoir} (c). This dynamics is defined as:
\begin{equation}
    P_o (k+1) = \mathcal{F}_r\left(P_o (k), P_i (k) \right) = \mathcal{F}_r\left(P_o (k), K_\text{in} \theta_d (k) \right),
\label{eq: nonlinear_mapping}
\end{equation}
where $\mathcal{F}_r(\cdot)$ denotes the nonlinear mapping with memory capacity (which is discussed in the next section) that governs the transformation from $P_i$ to $P_o$ through the dual-PAM actuator's high-dimensional structure. The passive PAM's inner pressure $P_o$ is employed as the readout of the physical reservoir's state.

\subsection{Hysteresis Characteristics}

\begin{figure}[tb]
    \centering
\includegraphics[width=\linewidth]{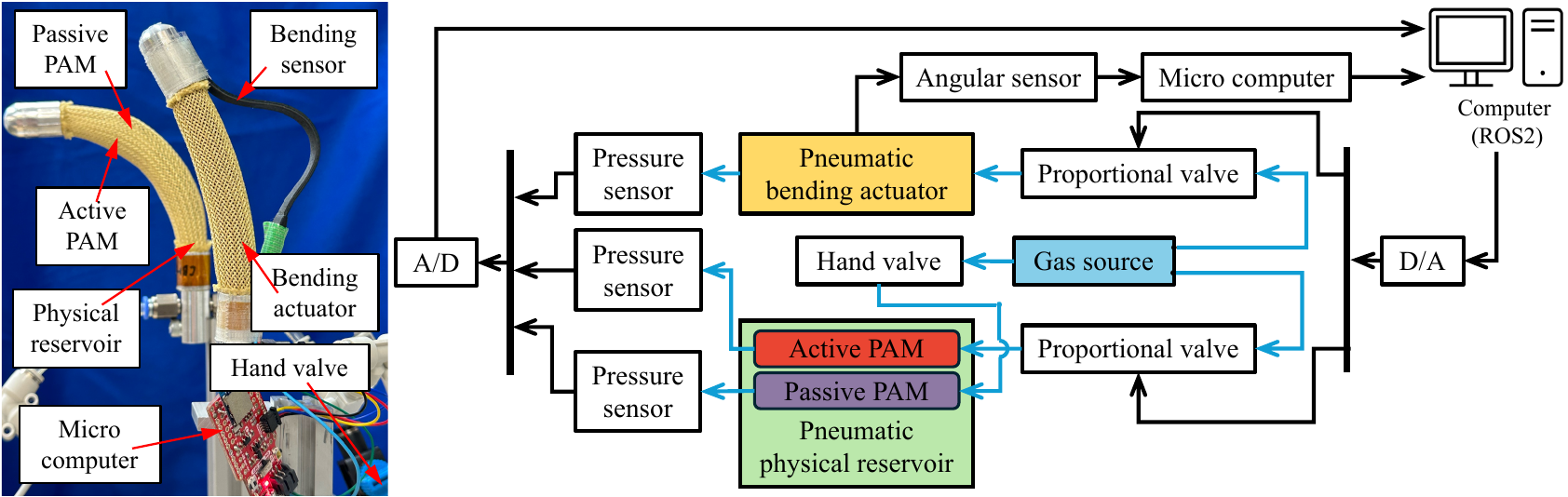}
    \caption{Experimental apparatus. Blue and black lines in the right-located figure indicate the pneumatic and electronic circuits, respectively.}
    \label{fig: exp_app}
    \vspace{-0.2cm}
\end{figure}

We used the experimental setup shown in Fig.~\ref{fig: exp_app} to obtain the hysteresis properties of the soft bending actuator and the physical reservoir. This setup comprises the soft actuator and the physical reservoir (both provided by Bridgestone), two proportional valves (Festo MPYE-5-M5-010B), an angular sensor (BENDLABS-1AXIS), three pressure sensors (SMC PSE540A-R06), and a computer with a Contec AI-1616L-LPE analog input board and a Contec AO-1608L-LPE analog output board. All experiment signal refresh rates were 200 Hz.

\begin{figure}[tb]
    \centering
\includegraphics[width=0.85\linewidth]{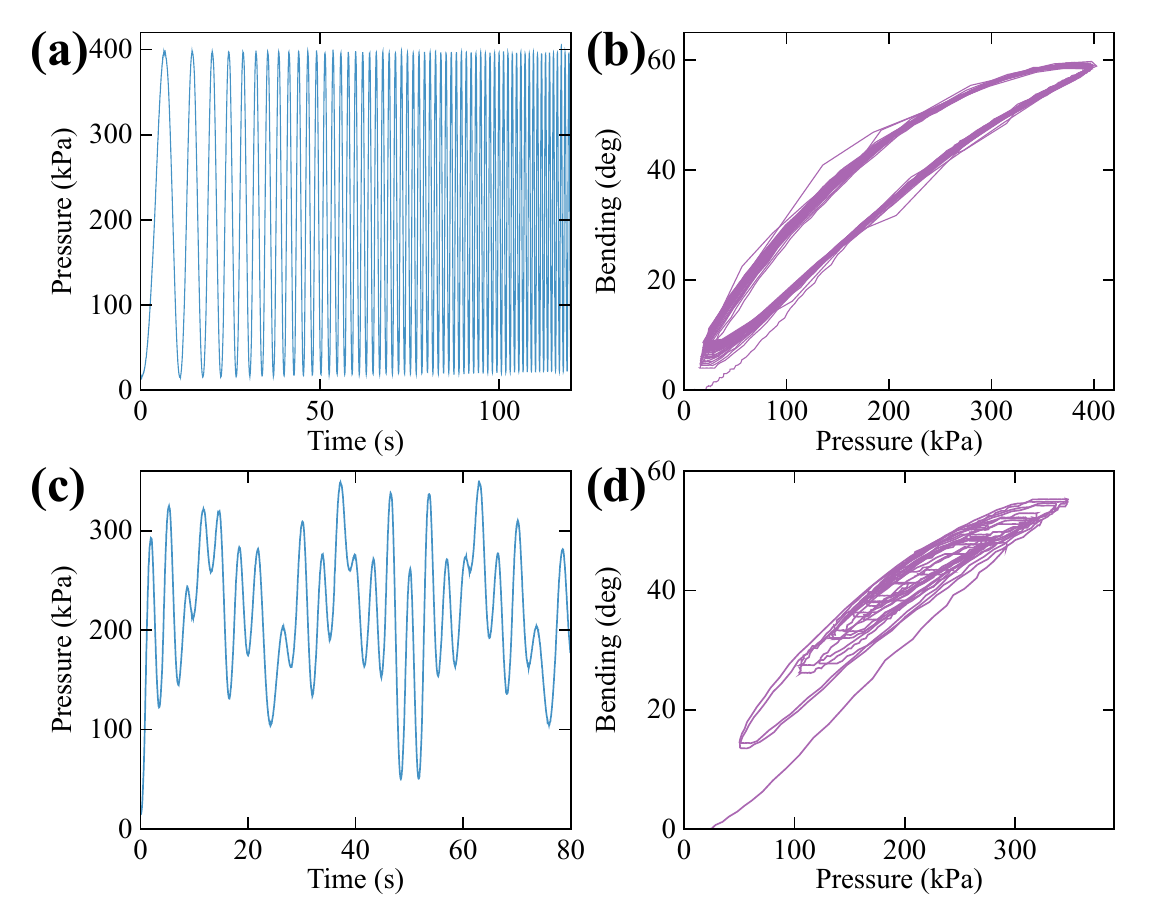}
    \caption{Hysteresis experiments: (a) and (b) Sweep-frequency pressure input $P_\text{exp}$ and the corresponding hysteresis loops of the first experiment; (c) and (d) Complex pressure input $P_\text{exp}$ and hysteresis loops of the second experiment. The signal in (a) lasts 120 [s] and sweeps the frequencies from 0.1 Hz to 1 Hz. The signal in (c) is $35\sum_{i=1}^{5} \sin\left(2\pi f_i t - \pi/2\right)+175$ [kPa], where $t \in [0,80]$ is the time [s], $f_1 = 0.12$, $f_2 = 0.04$, $f_3 = 0.31$, $f_4 = 0.29$, and $f_5 = 0.25$.}
    \label{fig: hys_loops}
    \vspace{-0.2cm}
\end{figure}

Two hysteresis experiments were conducted by pressurizing the bending actuator with different experiment pressure inputs $P_\text{exp}$ [kPa]: the first used a sweep-frequency signal shown by Fig.~\ref{fig: hys_loops} (a), and the second used a complex input shown by Fig.~\ref{fig: hys_loops} (c). During the experiments, the actuator's bending angle $\theta$ [deg] was fed into the physical reservoir by $P_i (k)= K_\text{in} \theta (k)$. $K_\text{in}=7.0$ was fixed throughout this study, this value was chosen based on the physical reservoir's actuation limit of 450 [kPa] and the bending actuator's output range of approximately 60 [deg]. We used the bending angle $\theta$ to generate $P_i$ at this stage to obtain the response of $P_o$ corresponding to $\theta$, facilitating the subsequent inverse operation in the development of the feedforward model. The pressure $P_i$ was regulated by a proportional valve and a feedback proportional-integral (PI) controller, whose gain settings will be detailed in Section IV.

\begin{figure}[tb]
    \centering
\includegraphics[width=0.85\linewidth]{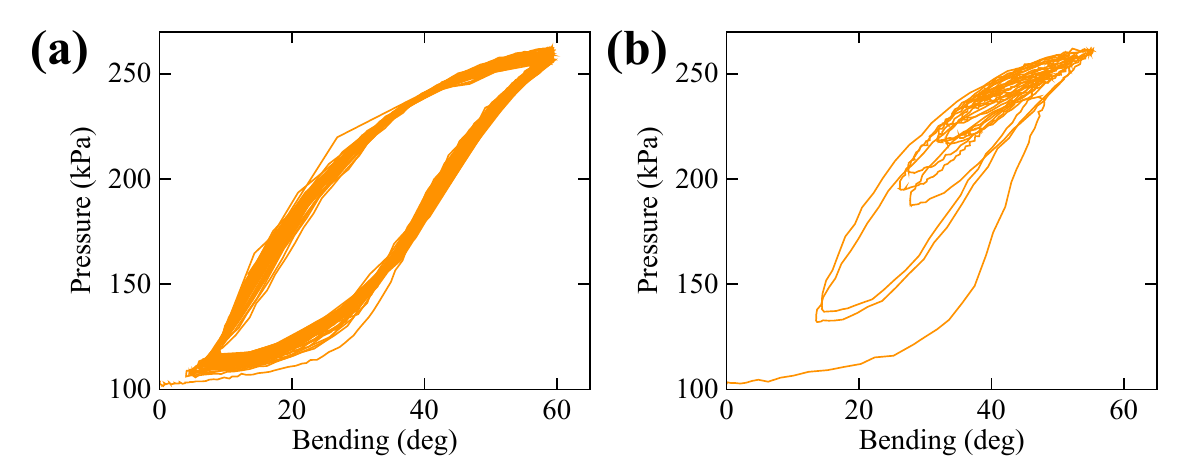}
    \caption{Hysteresis experiments: (a) Hysteresis between $P_o$ and the soft actuator's bending $\theta$ obtained from the first hysteresis experiment; (b) The corresponding hysteresis relationship obtained from the second experiment.}
    \label{fig: hys_reservoir}
    \vspace{-0.2cm}
\end{figure}

The hysteresis between $\theta$ and $P_\text{exp}$ obtained from the first and second experiments are depicted in Fig.~\ref{fig: hys_loops} (b) and (d), respectively. Similarly, the hysteresis between $P_o$ and $\theta$ during the two experiments are presented in Fig.~\ref{fig: hys_reservoir} (a) and (b). The hysteresis loops in both Fig.~\ref{fig: hys_loops} and Fig.~\ref{fig: hys_reservoir} are frequency-independent since the primary cause of the PAM-based soft actuators' hysteresis is the rate-independent Coulomb friction \cite{chou1996measurement}. As a time-dependent nonlinearity, the hysteresis observed in Fig.~\ref{fig: hys_loops} hinders the soft actuator's precise motion control  \cite{xie2018hysteresis}, while that shown in Fig.~\ref{fig: hys_reservoir} realizes the physical reservoir's nonlinear responses to input signals and its memory capacity, which is caused by the time-dependent properties of hysteresis.

\section{Feedforward Hysteresis Compensation}

In this section, we build two different feedforward hysteresis compensation models and conduct comparisons. The first model is based on a conventional ESN, and the second one is the T-S fuzzy PRC (FPRC) model proposed in this work.

\subsection{ESN Feedforward Hysteresis Compensation Model}

\begin{figure}[tb]
    \centering
\includegraphics[width=0.9\linewidth]{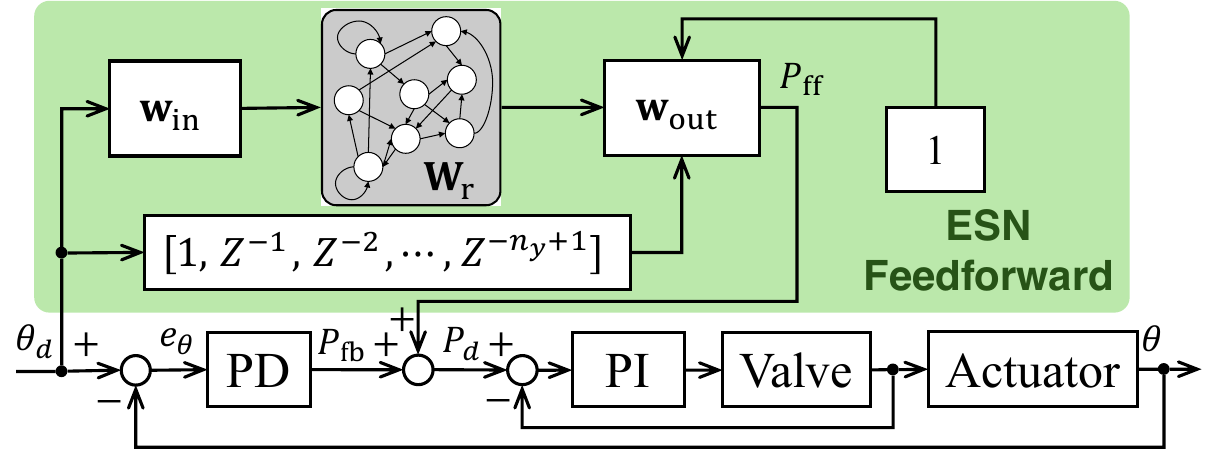}
    \caption{Control system with ESN feedforward hysteresis compensation.}
    \label{fig: esn system}
    \vspace{-0.2cm}
\end{figure}

Fig.~\ref{fig: esn system} is the schematic of the control system using the ESN feedforward model to control the soft actuator's bending motion. The system has a closed feedback loop and an open feedforward loop, where $e_\theta = \theta_d - \theta$ [deg] is the bending motion tracking control error, $P_\text{fb}$ [kPa] is the feedback adjustment from the proportional-differential (PD) controller in the closed loop, and $P_\text{ff}$ [kPa] is the feedforward hysteresis compensation model's output. The desired actuation pressure $P_d = P_\text{ff}+P_\text{fb}$ [kPa], is regulated by a PI pressure controller and used to pressurize the soft bending actuator. The PI pressure controller's performance is assumed to be ideal in this study.

Within the ESN feedforward model, $\mathbf{W}_r \in \mathbb{R}^{r \times r}$, the `reservoir', has a size of $r$, and the input layer $\mathbf{w}_\text{in} \in \mathbb{R}^r$ projects the input signal $\theta_d$ into a high-dimensional space, updating the hidden state $\mathbf{x} \in \mathbb{R}^r$ at each discrete time step as follows:
\begin{equation}
\mathbf{x}(k+1) = \gamma \ \mathbf{x}(k) + (1-\gamma) \tanh \left( \mathbf{w}_\text{in} \theta_d (k) + \mathbf{W}_r \mathbf{x}(k) \right),
\label{eq: esn state update}
\end{equation}
where $\gamma \in [0,1)$ is the leaky rate, and $\tanh(\cdot)$ is the hyperbolic tangent activation function commonly used in ESN models. The hidden state $\mathbf{x}$ is combined with the reference signal's temporal sequence $\bm{\theta}(k)=[\theta_d(k), \theta_d(k-1), \cdots, \theta_d(k-n_y+1)]^\top \in \mathbb{R}^{n_y}$ and a unit element 1 to form the extended state $\mathbf{x}^*(k) = [1, \bm{\theta}^\top(k), \mathbf{x}^\top(k)]^\top \in \mathbb{R}^{r+n_y+1}$. The extended state $\mathbf{x}^*$ is processed by the output layer $\mathbf{w}_\text{out} \in \mathbb{R}^{r+n_y+1}$ to produce:
\begin{equation}
P_\text{ff}(k) = \mathbf{w}_\text{out} \mathbf{x}^*(k) = \mathbf{w}_\text{out} [1, \bm{\theta}^\top(k), \mathbf{x}^\top(k)]^\top,
\label{eq: P_ff}
\end{equation}
where $P_\text{ff}$ is the feedforward model's output corresponding to the given desired bending angle $\theta_d$. $\mathbf{w}_\text{in}$ and $\mathbf{W}_r$ are initially randomized with parameters from a $N(0,1)$ distribution and subsequently fixed after being adjusted by the scaling factor $f_\text{input}$, which scales the input weights in $\mathbf{w}_\text{in}$, and the spectral radius $\rho$, which adjusts the largest absolute eigenvalue of $\mathbf{W}_r$. The only trainable parameters in the model are linear weights in $\mathbf{w}_\text{out}$, which are optimized using the linear regression with Tikhonov regularization (ridge regression) as expressed by:
\begin{equation}
\mathbf{w}_\text{out} = \argmin_{\mathbf{w}_\text{out}} \left(\sum_{k \in K_{\text{train}}}\left(P_\text{exp}(k) - \mathbf{w}_\text{out} \mathbf{x}^*(k) \right)^2 + \alpha \left\lVert \mathbf{w}_\text{out} \right\rVert^2 \right),
\label{eq: esn ridge}
\end{equation}
where $K_\text{train}$ is to the training data obtained from the aforementioned first hysteresis experiment. $\left\rVert \cdot \right\rVert$ denotes the L2 norm. $\alpha$ is the regularization factor. The key hyperparameters of the ESN model include the reservoir size $r$, the leaky rate $\gamma$, the input weight scaling factor $f_\text{input}$, and the spectral radius $\rho$. 

The dataset $K_\text{train}$ comprises $P_\text{exp}$, $P_o$, and $\theta$. The objective of training the feedforward model is to obtain an approximation of the inverse projection depicted in Fig.~\ref{fig: hys_loops} that can estimate the actuator's input $P_\text{exp}$ given the output $\theta$ (and its nonlinear transformation performed by the ESN or the PRC). Hence, the experimentally obtained $\theta$ serves as the desired $\theta_d$ in (\ref{eq: convert}), (\ref{eq: esn state update}), and (\ref{eq: P_ff}) in the model training and test phases. The training process seeks to satisfy $P_\text{ff} \approx P_\text{exp}$ by the optimization shown in (\ref{eq: esn ridge}) with $P_\text{exp}$ shown in Fig.~\ref{fig: hys_loops} (a) and (c) as the ground truth.

\subsection{Fuzzy PRC Feedforward Hysteresis Compensation Model}

\begin{figure}[tb]
    \centering
\includegraphics[width=\linewidth]{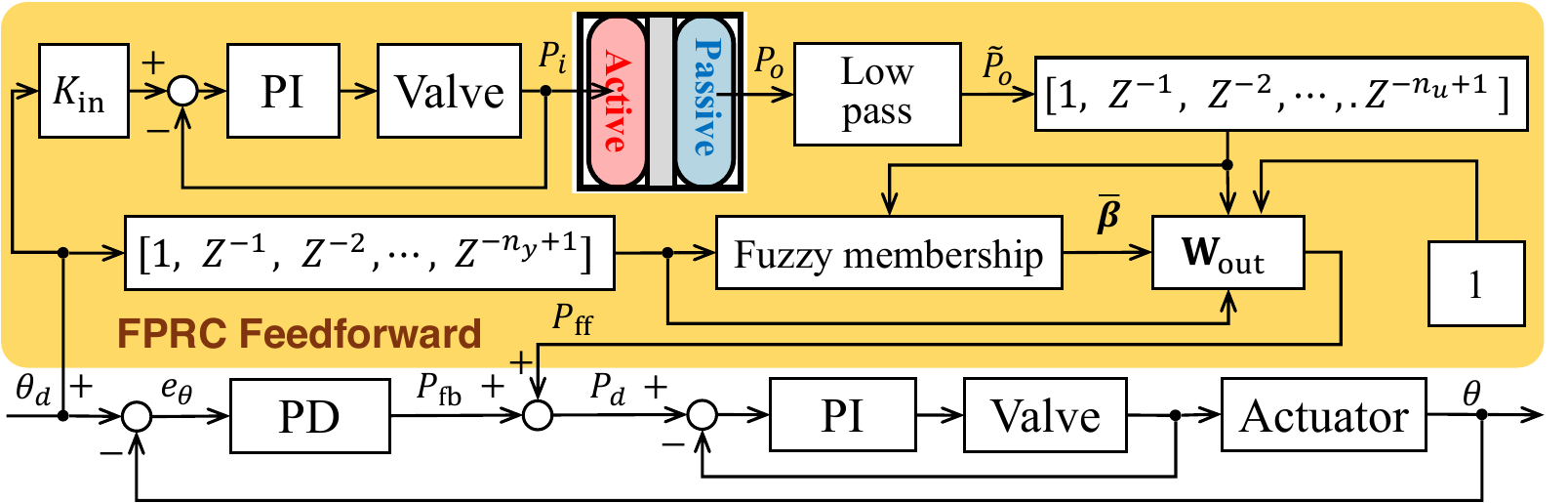}
    \caption{Control system with FPRC feedforward hysteresis compensation.}
    \label{fig: prc system}
    \vspace{-0.2cm}
\end{figure}

Fig.~\ref{fig: prc system} presents a schematic of the control system with the FPRC feedforward model, where the closed-loop components are consistent with Fig.~\ref{fig: esn system} with the reservoir part $\mathbf{W}_r$ in the ESN being replaced by the physical reservoir in Fig.~\ref{fig: reservoir}. The FPRC model uses $K_\text{in}$ to convert $\theta_d$ into $P_i$ by (\ref{eq: convert}). $P_i$ is regulated by a PI pressure controller, whose performance is assumed ideal. 

It is essential to clarify that while the performance of the PI pressure controllers is deemed ideal in both the closed-loop control system and the FPRC model, these assumptions are based on different reasons. The PI controller is considered ideal in the closed-loop control system as the pressure control task is relatively simple and typically performs satisfactorily compared to controlling the soft actuator's bending motion. Therefore, the pressure control is not the primary focus of this research. Conversely, the assumption of ideal performance for the PI controller in the FPRC model is based on the fact that the proportional valve, shown in Fig.~\ref{fig: exp_app} and controlled by the PI controller, also has its dynamics. The valve's dynamics and any pressure control errors are integrated into the overall dynamics of the nonlinear physical system in the FPRC model. Thus, the pressure control errors do not require consideration.

The ESN model's leaky rate $\gamma$ is mimicked in the FPRC model by a low-pass filter with the filter factor $\epsilon \in (0,1]$ as:
\begin{equation}
\tilde{P_o}(k) = \epsilon P_o(k) + (1-\epsilon) \tilde{P_o}(k-1),
\end{equation}
where $\tilde{P_o}$ is the filtered state of the physical reservoir. Unlike the state $\mathbf{x}$ in the ESN, which can be defined with arbitrary dimensionality, the physical reservoir's state is represented by a single scalar $\tilde{P_o}$. Due to the absence of explicitly tunable hyperparameters, the adjustability of the physical reservoir's computational performance is comparatively limited \cite{lee2024task}. Previous works used the time sequences of the physical reservoir's readouts for computation \cite{hayashi2022, akashi2024embedding}. Similarly, we collect the physical reservoir's outputs into temporal sequences of size $n_u$, as $\mathbf{p}(k) = [\tilde{P_o}(k), \tilde{P_o}(k-1), \cdots, \tilde{P_o}(k-n_u+1)]^\top \in \mathbb{R}^{n_u}$. The FPRC model's properties can then be adjusted by altering $n_u$.

Motivated by the precise hysteresis modeling achieved by using the T-S fuzzy model to process time sequences \cite{cheng2016adaptive}, we integrate a T-S fuzzy logic into the FPRC model. The T-S fuzzy model begins with clustering the extended state vector $\mathbf{x}^\dag(k) = [\bm{\theta}^\top(k), \mathbf{p}^\top(k)]^\top \in \mathbb{R}^{n_y+n_u}, \ k \in K_\text{train}$ into $n_c$ fuzzy clusters by the Fuzzy C-Means (FCM) algorithm \cite{bezdek1984fcm}, where $n_c$ is the number of T-S fuzzy rules. In each fuzzy cluster $i, \ i \in \{1, 2, \cdots, n_c\}$, the state $\mathbf{x}^\dag (k)$ has a membership $u_i(k) \in (0,1) \ s.t. \sum_1^{n_c}u_i(k) = 1$, and is processed by an output layer $\mathbf{w}^i_\text{out} \in \mathbb{R}^{n_y+n_u+1}$ to produce an output $P^i_\text{ff}(k)$ as follows:
\begin{equation}
P^i_\text{ff}(k) = \mathbf{w}^i_\text{out}\left[1, {\mathbf{x}^\dag}^\top(k)\right]^\top = \mathbf{w}^i_\text{out} \left[1, \bm{\theta}^\top(k), \mathbf{p}^\top(k)\right]^\top.
\label{eq: fuzzy ff}
\end{equation}
The output layers, $\mathbf{w}^i_\text{out}, \cdots, \mathbf{w}^{n_c}_\text{out}$, are separately learned by:
\begin{equation}
\mathbf{w}^i_\text{out} = \argmin_{\mathbf{w}^i_\text{out}} \left( \sum_{k \in K_\text{train}} u_i^2(k)\left( P_\text{exp}(k) - P^i_\text{ff}(k) \right)^2 + \alpha \left\lVert \mathbf{w}^i_\text{out} \right\rVert^2 \right),
\end{equation}
and combined into an output matrix $\mathbf{W}_\text{out} \in \mathbb{R}^{n_c \times (n_y+n_u+1)}$ as:
\begin{equation}
\mathbf{W}_\text{out} = \left[{\mathbf{w}^1}_\text{out}^\top, {\mathbf{w}^2}_\text{out}^\top, \cdots, {\mathbf{w}^{n_c}}_\text{out}^\top\right]^\top.
\end{equation}
During execution (including the model validation and test) the membership $\beta_i \in (0,1)$ of the state $\mathbf{x}^\dag$ to the $i$-th fuzzy cluster is computed using the Gaussian membership function as follows:
\begin{equation}
\beta_i(k) = \exp \left( -\frac{1}{2\sigma^2} \| \mathbf{x}^\dag(k) - \mathbf{c}_i \|^2 \right),
\end{equation}
where $\sigma$ is a scaling factor and $\mathbf{c}_i \in \mathbb{R}^{n_y+n_u}$ is the center of the $i$-th cluster obtained at the initial clustering using FCM. The obtained memberships are normalized into $\Bar{\bm{\beta}} \in \mathbb{R}^{n_c}$ as:
\begin{equation}
\Bar{\bm{\beta}}(k) = \frac{1}{\sum_{i=1}^{n_c}\beta_i(k)} \left[\beta_1(k), \beta_2(k), \cdots, \beta_{n_c}(k)\right],
\end{equation}
which is then used to compute the FPRC model's output by:
\begin{equation}
P_\text{ff}(k) = \Bar{\bm{\beta}}(k) \mathbf{W}_\text{out} \left[1, {\mathbf{x}^\dag}^\top(k)\right]^\top = \Bar{\bm{\beta}}(k) \mathbf{W}_\text{out} \left[\bm{\theta}^\top(k), \mathbf{p}^\top(k)\right]^\top.
\label{eq: fuzzy output}
\end{equation}

\subsection{Qualitative Model Comparison}

The FPRC model offers several advantages over the ESN model regarding the implementation as the feedforward hysteresis compensation model: 1) The FPRC model leverages the inherent nonlinear properties of physical systems, precisely, the dual-PAM soft actuator in this work, to model hysteresis. Hence, it is less computationally expensive in real-time control applications than the ESN, as the real-world nonlinear physical system performs the state update \cite{akashi2024embedding}. 2) The FPRC model relies on a real-world physical system for computation, where well-understood physical laws, e.g., energy conservation, govern the physical system's dynamical behaviors. Consequently, the physical system's responses to specific input signals are usually monotonic and thus intuitively more understandable from a real-world view than the ESN's complex state variations that may contain more than tens of dimensions \cite{park2016online}.

Nevertheless, the FPRC model also presents several disadvantages: 1) The model's reliance on an actually existing physical system introduces additional requirements for devices and space, such as the dual-PAM actuator, proportional valves, and pressure sensors used in this work, which potentially increases the system's overall bulkiness. 2) Unlike the ESN, where performance can be easily tuned by hyperparameters, the computational capacity of the physical system is difficult to adjust and analyze \cite{lee2024task}. Additional techniques, e.g., the temporal sequence processing and the T-S fuzzy model employed in this work, may be necessary to compensate for this lack of flexibility in the PRC's performance. These techniques may require domain expertise and introduce additional computational demands. 3) The exact mathematical description of the nonlinear dynamics, which is explicit in the ESN, can often be challenging to ascertain from real-world physical systems.

\subsection{Quantitative Model Comparison}

To quantitatively assess the ESN and FPRC models, we used the data shown in Fig.~\ref{fig: hys_loops} (a), (b) and Fig.~\ref{fig: hys_reservoir} (a) for model training, and used the data from Fig.~\ref{fig: hys_loops} (c), (d) and Fig.~\ref{fig: hys_reservoir} (b) to test the trained model. A comparative linear model (named Fuzzy linear) was constructed by substituting $[1, {\mathbf{x}^\dag}^\top(k)]^\top$ in (\ref{eq: fuzzy ff}) and (\ref{eq: fuzzy output}) with $[1, \bm{\theta}^\top(k)]^\top$ to illustrate the contribution of the physical reservoir's readout $\tilde{P_o}$ within the FPRC model.

Table~\ref{tab: model_config} lists the parameters for ESN and FPRC models. Before model training and test, a 100-step washout was used to initialize the ESN's state $\mathbf{x}$. The number of fuzzy rules $n_c$ in the comparative Fuzzy linear model was set consistent with the FPRC model. All models (i.e., FPRC, ESN, and Fuzzy linear) have the tap size $n_y$ of 5 and the ridge parameter $\alpha$ of 0.001. The parameters for the FPRC model were chosen by trial and error, while those for the ESN were selected to align its performance with the FPRC model in the training stage.

\begin{table}[tb]
\centering
\caption{Configuration Parameters for Different Models}
\label{tab: model_config}
\renewcommand{\arraystretch}{0.5}
\begin{tabular}{lll}
\toprule
\textbf{Model}        & \textbf{Parameter}                      & \textbf{Value}   \\
\midrule
\multirow{4}{*}{ESN}  & Reservoir size, $r$                     & 800             \\
                      & Input scaling factor, $f_\text{input}$  & 0.02            \\
                      & Leaky rate, $\gamma$                    & 0.8             \\
                      & Spectral radius, $\rho$                 & 0.4             \\
\midrule
\multirow{4}{*}{FPRC} & Filter factor, $\epsilon$               & 0.01            \\
                      & Fuzzy clusters, $n_c$                   & 8               \\
                      & Reservoir output tap size, $n_u$        & 3               \\
                      & Gaussian membership scaling, $\sigma$   & 2.0               \\
\bottomrule
\end{tabular}
\vspace{-0.2cm}
\end{table}

All models were trained by a 5-fold cross-validation process, where each model was independently trained five times, each time using differently selected 80\% of the training set for training and the remaining 20\% for validation. The root-mean-square errors (RMSE) of the training ($e_\text{train}$) and the validation ($e_\text{val}$) were recorded during the cross-validation. The model yielding the minimal normalized error $\Bar{e} = \sqrt{e^2_\text{train}+e^2_\text{val}}$ was deemed the optimal training outcome and subsequently evaluated on the test dataset. The training and test procedures were repeated ten times to compute the execution time costs.

\begin{figure}[tb]
    \centering
\includegraphics[width=0.9\linewidth]{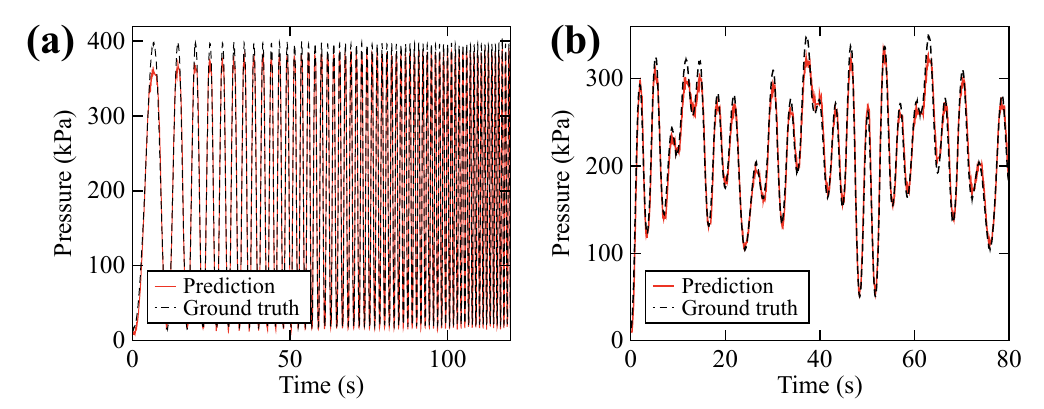}
    \caption{Training and test results of the FPRC model: (a) Training; (b) Test.}
    \label{fig: training}
\end{figure}

Demonstrative training and test results of the FPRC model are presented in Fig.~\ref{fig: training}. The average and standard deviation of RMSE values during the 5-fold cross-validation, the RMSE values in the test phase, and the average and standard deviation of execution time are shown in Table~\ref{tab: training numerical}, where bold font highlights the best-performing model for each comparison metric. As shown in Fig.~\ref{fig: training}, the FPRC model's results align closely with the ground truth $P_\text{exp}$ in both the training and the test phase. Numerically, despite the similarities in training and validation performance between the ESN and FPRC models, the FPRC model demonstrates a lower RMSE in the test phase. The FPRC model's test RMSE is consistent with its training outcomes, whereas the ESN model exhibits higher test RMSE values compared to its training and validation results. The RMSE comparison between the FPRC model and the Fuzzy linear model, where the only difference between them is the inclusion of $\tilde{P_o}$, highlights the contribution of the physical reservoir to the FPRC model's improved accuracy. However, due to the additional features ($\tilde{P_o}$) included, the FPRC model exhibits longer execution times than the Fuzzy linear model. Compared to the ESN model, the FPRC model required less execution time for its training. Since the ESN's state update, as (\ref{eq: esn state update}) shows, involves the operation of a large matrix $\mathbf{W}_r$, which is instead processed by the physical system in the FPRC model, the FPRC model achieves a 97.6\% reduction in execution time during the test phase than the ESN model.

\begin{table}[tb]
\centering
\caption{Numerical comparison among different models.}
\label{tab: training numerical}
\setlength{\tabcolsep}{1.5pt}
\renewcommand{\arraystretch}{0.6}
\begin{tabular}{lccc}
\toprule
{Performance Matrix} & {ESN} & {FPRC} & {Fuzzy Linear} \\
\midrule
Training RMSE [kPa]    & \textbf{11.5$\pm$0.0143} & 11.6$\pm$0.188  & 14.9$\pm$0.0167 \\
Validation RMSE [kPa] & \textbf{11.5$\pm$0.0479} & 11.7$\pm$0.154  & 14.9$\pm$0.0587 \\
Test RMSE [kPa]  & 16.5 & \textbf{11.0} & 16.7 \\
Training time cost [s] & 16.6$\pm$3.28 & 12.0$\pm$2.39 & \textbf{8.98$\pm$0.994} \\
Test time cost [ms]  & 8930$\pm$1800 & 210$\pm$4.36 & \textbf{191$\pm$9.61} \\
\bottomrule
\end{tabular}
\vspace{-0.2cm}
\end{table}

We compared the three models by reversing the original training and test data for test and training (only here). Table \ref{tab: reverse training numerical} shows the results. Since the original training data contains more high-frequency signals than the original test data (as Fig.~\ref{fig: training} shows), all models exhibit worse test RMSE than that in Table \ref{tab: training numerical} due to unseen frequency features met at the test phase. The FPRC exhibits the lowest RMSEs among all models, highlighting its better generalization with unseen information.

\begin{table}[t]
\centering
\caption{Comparison among models with reversed data set.}
\label{tab: reverse training numerical}
\setlength{\tabcolsep}{1.2pt}
\renewcommand{\arraystretch}{0.8}
\begin{tabular}{lccc}
\toprule
{Performance Matrix} & {ESN} & {FPRC} & {Fuzzy Linear} \\
\midrule
Training RMSE [kPa]    & 10.2$\pm$0.0341 & \textbf{9.07$\pm$0.0164}  & 13.5$\pm$0.0668 \\
Validation RMSE [kPa] & 10.2$\pm$0.142 & \textbf{9.07$\pm$0.148}  & 13.5$\pm$0.114 \\
Test RMSE [kPa]  & 60.32 & \textbf{26.2} & 51.5 \\
\bottomrule
\end{tabular}
\vspace{-0.2cm}
\end{table}

\begin{figure}[tb]
    \centering
\includegraphics[width=0.85\linewidth]{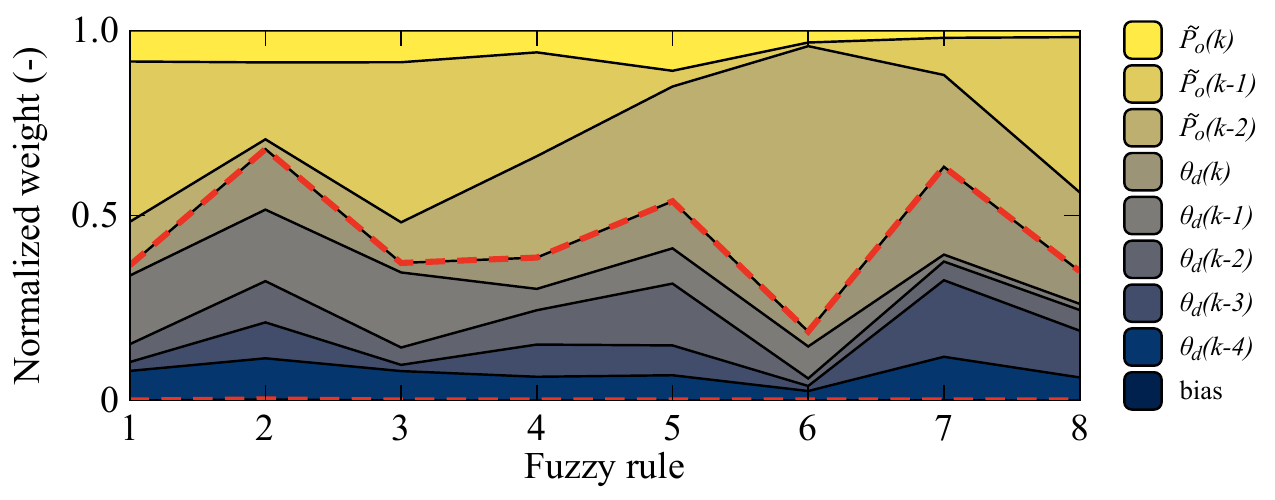}
    \caption{Proportional contribution of normalized weights. The upper dashed red line is the separation between the weights corresponding to $\tilde{P_o}$ and $\theta_d$, the lower dashed red line is the separation between $\tilde{P_o}$ and the bias term.}
    \label{fig: portion}
    \vspace{-0.2cm}
\end{figure}

The greater magnitude of a linear weight can indicate its corresponding nonlinear feature's more predominant position in describing the nonlinear dynamics \cite{brunton2016discovering}. To quantitatively clarify the physical reservoir's contribution within the FPRC model, we first normalized the training data by the following:
\begin{equation}
\Bar{x}(k) = \frac{x(k) - \min(x)}{\max(x) - \min(x)},
\end{equation}
where $x$ represents either $\tilde{P_o}$ or $\theta$ (used as $\theta_d$ in training), and $\min(x)$ ($\max(x)$) denotes the minimum (maximum) value of $x$ within the training dataset. Then, the normalized data were used to train the FPRC model by the aforementioned process, and the obtained linear weights were normalized as follows:
\begin{equation}
\bar{\omega}_j = \frac{|\omega_j|}{\sum_{i=1}^{n_u+n_y+1}|\omega_i|},
\end{equation}
where $|\cdot|$ means the absolute value, and $\omega_j, \ j \in \{1, 2, \cdots, n_u+n_y+1\}$ denotes the individual linear weight (including bias) contained in $\mathbf{w}^q_\text{out}$ of each fuzzy rule $q$, $q\in\{1, 2, \cdots, n_c\}$. The normalized weights corresponding to different nonlinear features of the FPRC model, including $\tilde{P_o}$, $\theta_d$, and the bias term, are shown in Fig.~\ref{fig: portion}. As shown in Fig.~\ref{fig: portion}, the importance of different features varies among different fuzzy rules. The upper dashed red line highlights that the weights combining $\tilde{P_o}$ exhibit the same magnitude as those processing $\theta_d$. On average, the weights corresponding to $\tilde{P_o}$ and $\theta_d$ account for 56.1\% and 43.8\% across the eight fuzzy rules, respectively.

\subsection{Performance Analysis Based on Control Variable}

\begin{figure}[tb]
    \centering
\includegraphics[width=\linewidth]{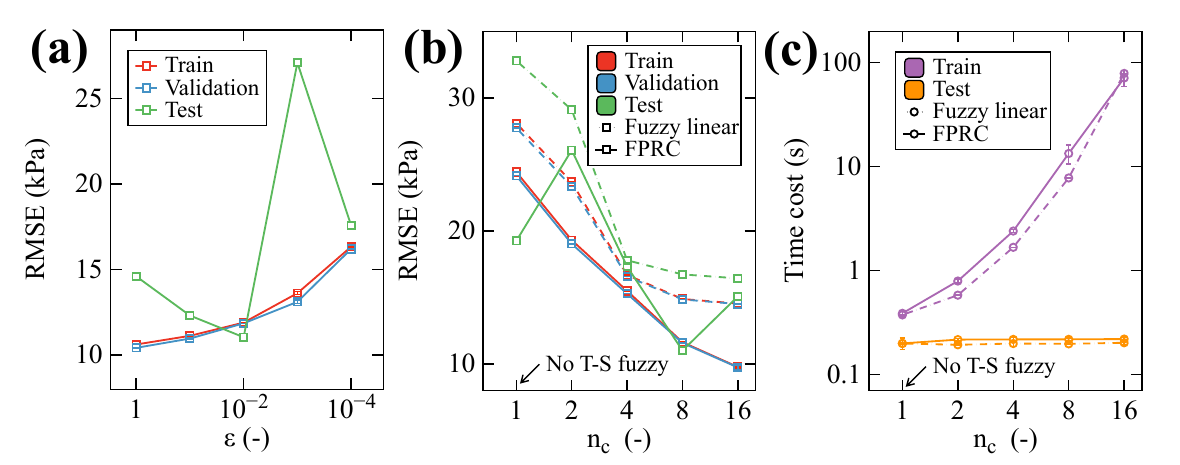}
    \caption{ Training and test results: (a) Variations of RMSEs with $\epsilon$ values; (b) and (c) Variations of RMSEs and execution time costs with $n_c$ values.}
    \label{fig: statistic}
    \vspace{-0.2cm}
\end{figure}

We used the control variable method to analyze the influence of different parameters on the FPRC model's performance. Specifically, we used the filter factor $\epsilon$ being 1, 0.1, 0.01, 10$^{-3}$, and 10$^{-4}$ to train the FPRC model with all other parameters consistent with the previous section. Similarly, we trained the FPRC and the Fuzzy linear models with $n_c$ being 1, 2, 4, 8, and 16, with all other parameters fixed. The results are presented in Fig.~\ref{fig: statistic}. The results show that the increase in $\epsilon$ and $n_c$ causes the FRPC model's training RMSE to decrease, while the test RMSE exhibits complex non-monotonic variations. Results in (c) show that greater $n_c$ leads to longer training time as a larger group of weights are optimized, but the test time cost barely changes. Despite the longer execution times, the FPRC model consistently shows lower training, validation, and test RMSE values than the Fuzzy linear model across the tested settings.

\begin{figure}[tb]
    \centering
\includegraphics[width=\linewidth]{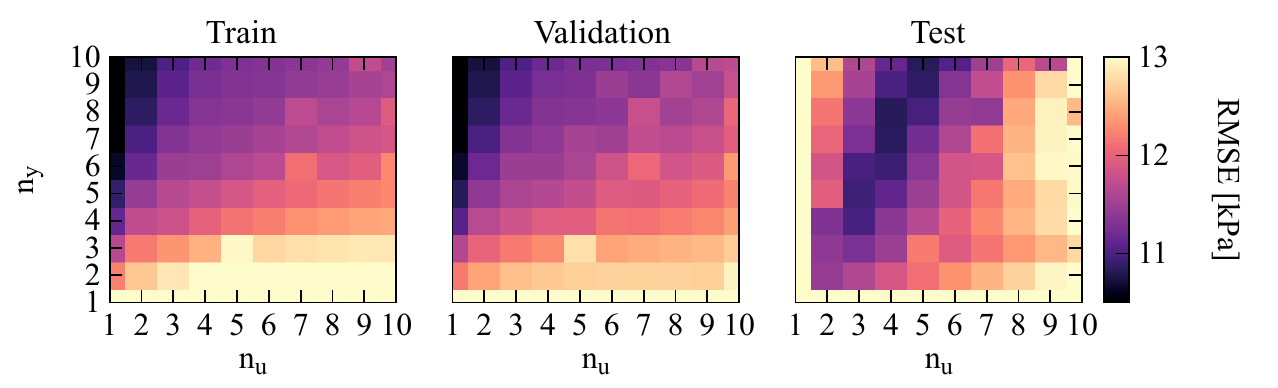}
    \caption{Model errors with different tap sizes, including RMSEs in training, validation, and test stages. Results were obtained with $n_c=8$, and $\epsilon=0.01$.}
    \label{fig: tap_sizes}
    \vspace{-0.2cm}
\end{figure}

The previous section has shown the necessity of $\tilde{P_o}$ within the FPRC model, i.e., $n_u>0$. We extended this analysis by training the FPRC model with different $n_u$ and $n_y$ (from 1 to 10) combinations with all other parameters fixed, and the results are shown in Fig.~\ref{fig: tap_sizes}. It is observed that while the model's training and validation performance benefits from less $n_u$ and greater $n_y$, a medium $n_u$ selection helps decrease the RMSE value in the model test stage, benefiting its generalization.

Theoretically explaining the test performance's variation with model parameters, e.g., the `double descent' phenomenon \cite{nakkiran2020optimal}, is challenging and needs domain expertise. As our scope is developing the FPRC model to control soft actuators, further analysis of the test error's complex variation is omitted here.

\section{Experimental Results and Discussion}

We tested the above-trained FPRC model alone and with a PD controller (FPRC+PD) in controlling the soft actuator's bending angle tracking different reference signals. The control system's structure aligns with Fig.~\ref{fig: prc system}. The feedback controller gains were selected by trial and error. The P and D gains of the PD controller in the main feedback loop were 0.5 [kPa/deg] and 0.005 [kPa$\cdot$s/deg], the P and I gains of the PI controller in the main loop were 0.03 [V/kPa] and $0.1\times10^{-5}$ [V/kPa$
\cdot$s], the P and I gains of the PI controller within the FPRC model were 0.05 [V/kPa], and $0.1\times10^{-5}$ [V/kPa$\cdot$s], respectively.

\begin{figure}[tb]
    \centering
\includegraphics[width=0.8\linewidth]{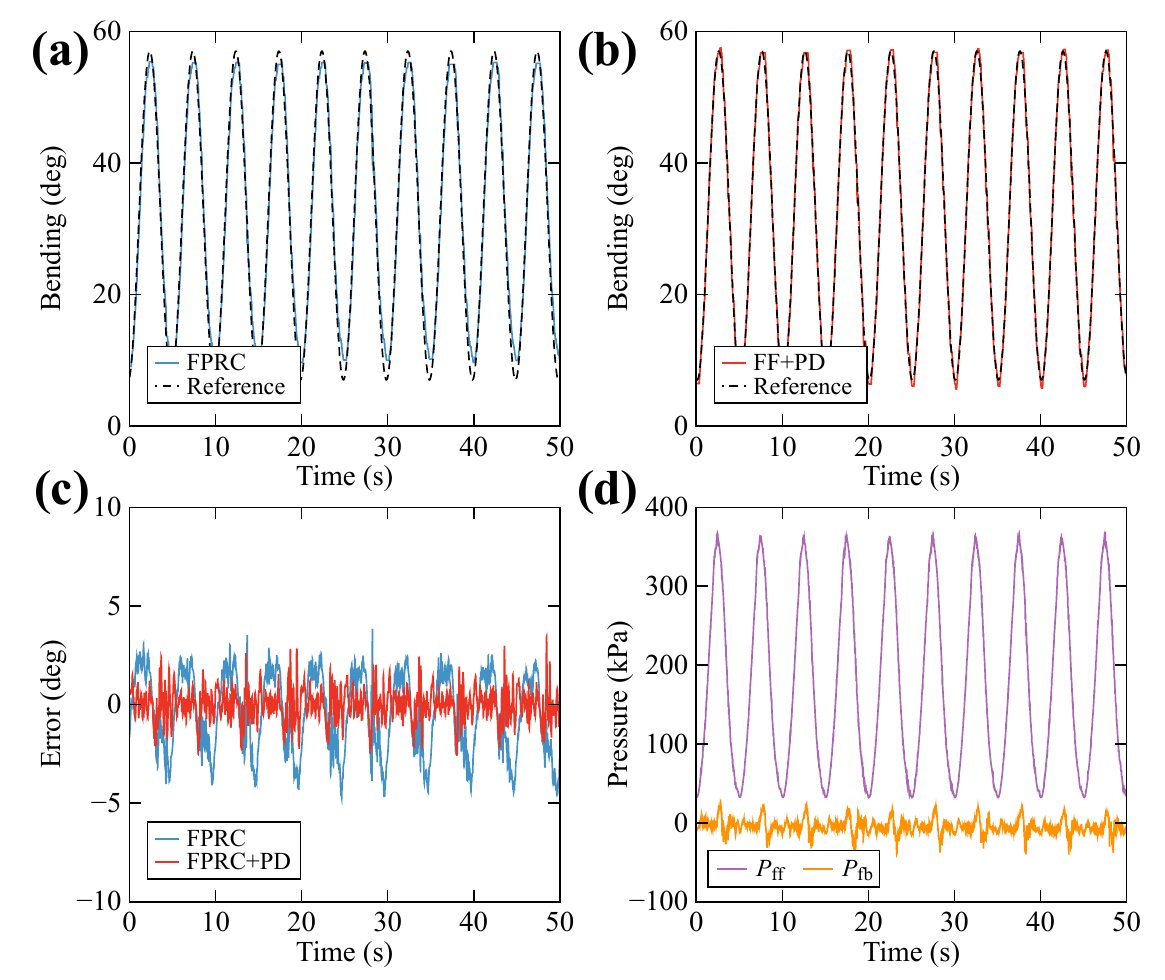}
    \caption{Tracking a 0.2Hz sine wave: (a) Open-loop control; (b) Closed-loop control; (c) Control errors; (d) Feedforward and feedback controller outputs.}
    \label{fig: 0.2}
    \vspace{-0.2cm}
\end{figure}

\begin{figure}[tb]
    \centering
\includegraphics[width=0.8\linewidth]{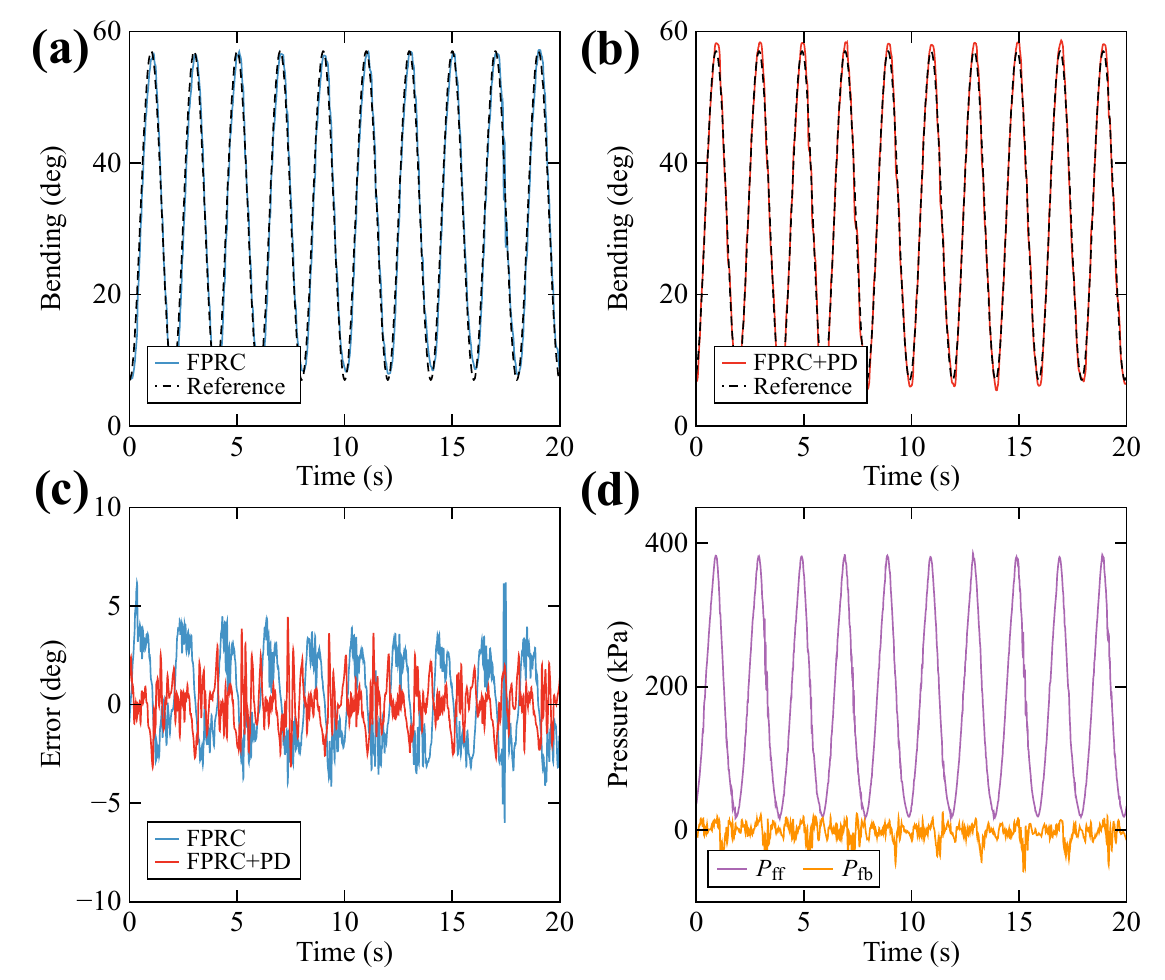}
    \caption{Tracking a 0.5Hz sine wave: (a) Open-loop control; (b) Closed-loop control; (c) Control errors; (d) Feedforward and feedback controller outputs.}
    \label{fig: 0.5}
    \vspace{-0.2cm}
\end{figure}

Experimental results of tracking 0.2 Hz and 0.5 Hz sine waves are shown in Fig.~\ref{fig: 0.2} and Fig.~\ref{fig: 0.5}. The results indicate that both the open-loop control with the standalone FPRC model and the closed-loop FPRC+PD control system closely followed the reference signals. The closed-loop control method performs better because of the feedback controller's adjustment capabilities. Variations of the feedforward model's output $P_\text{ff}$ and the feedback adjustments $P_\text{fb}$ during experiments are shown in Fig.~\ref{fig: 0.2}(d) and Fig.~\ref{fig: 0.5}(d). These results show that the FPRC model's output $P_\text{ff}$ is the predominant element in the FPRC+PD closed-loop control system, with the PD controller's output $P_\text{fb}$ primarily serving as fine adjustments.

\begin{figure}[tb]
    \centering
\includegraphics[width=0.8\linewidth]{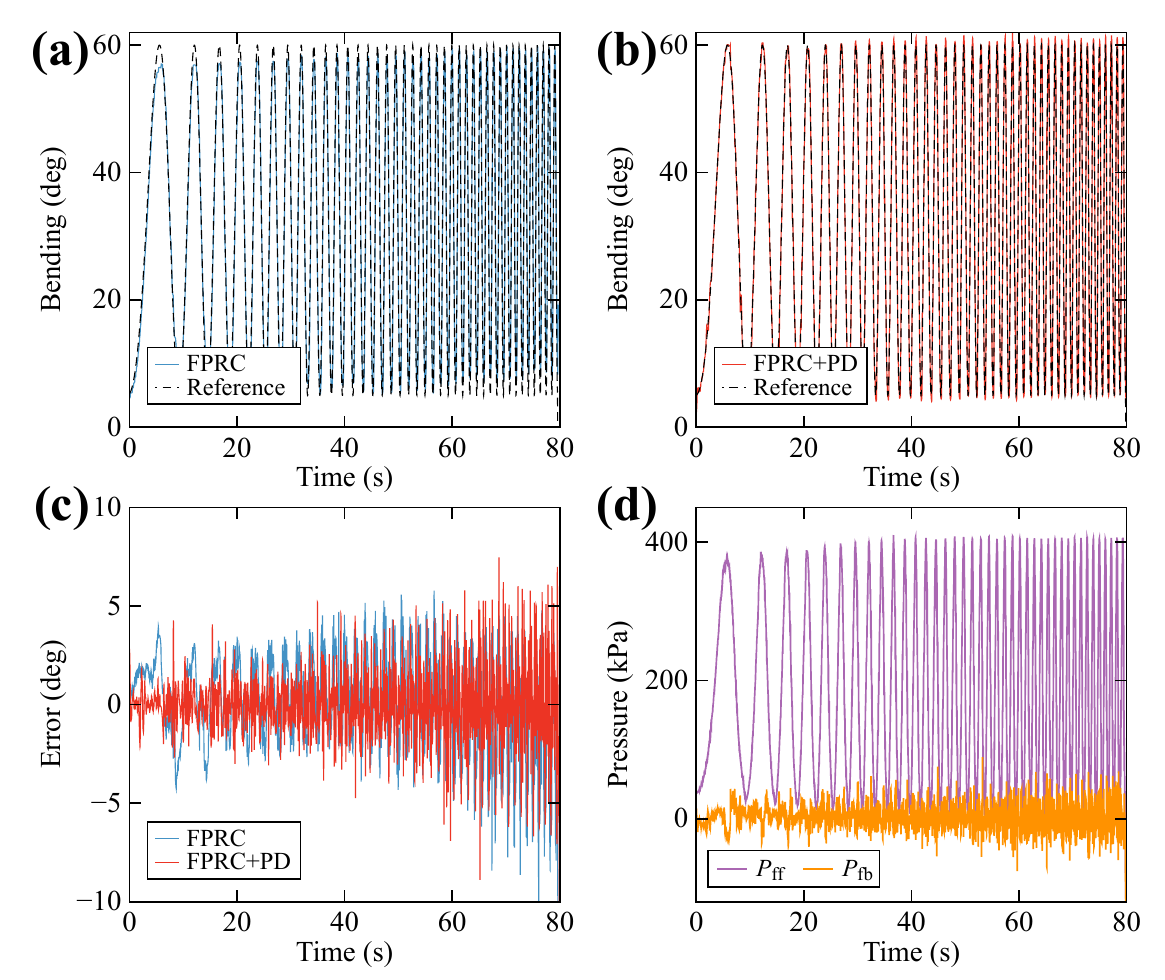}
    \caption{Tracking a sweep-frequency reference signal: (a) Open-loop control; (b) Closed-loop control; (c) Control errors; (d) Feedforward and feedback controller outputs. The reference bending angle is expressed as $27.5 \sin\left(\pi t (0.01125 t+0.1)-0.5\pi\right) + 32.5$ [deg], $t \in [0,80]$ is the time [s].}
    \label{fig: sweep}
    \vspace{-0.2cm}
\end{figure}

\begin{figure}[tb]
    \centering
\includegraphics[width=0.8\linewidth]{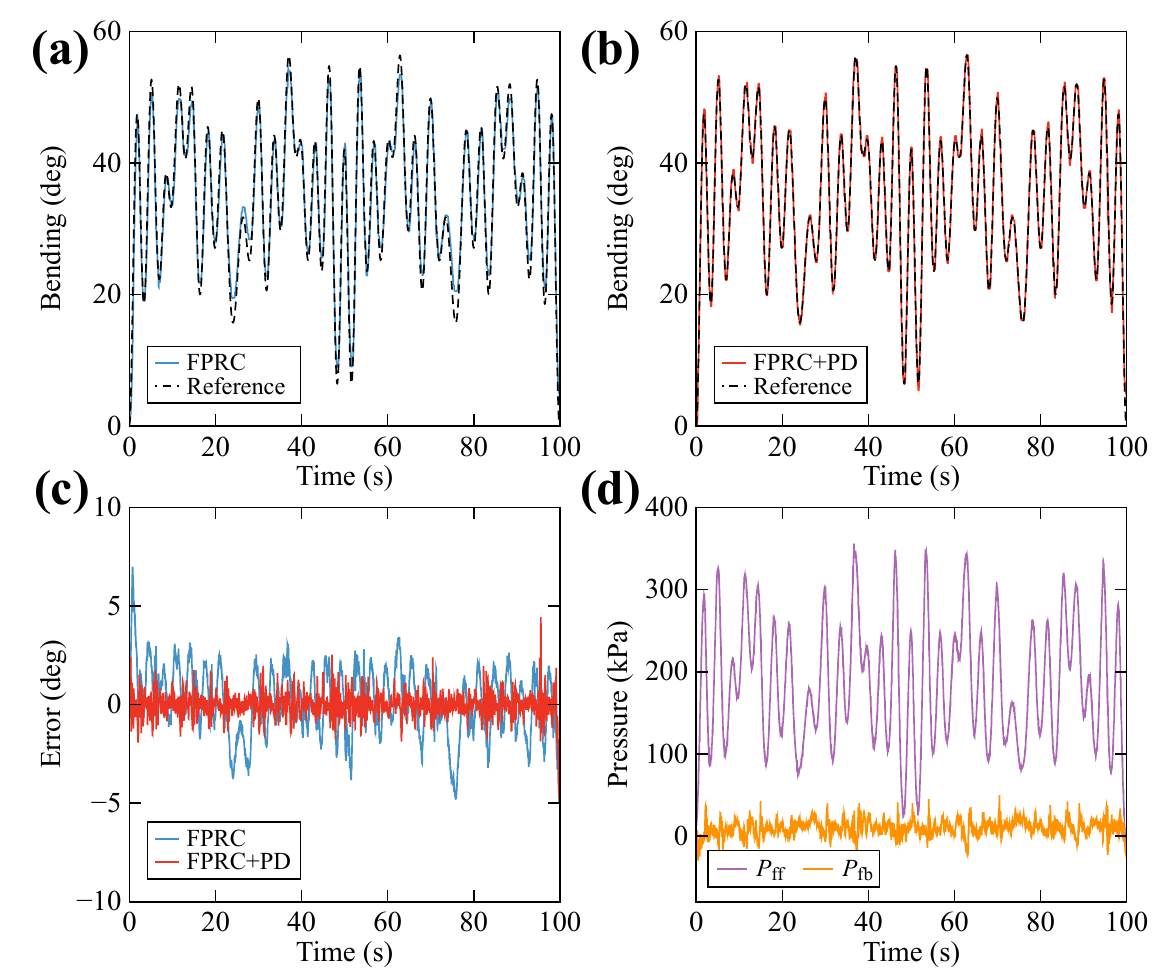}
    \caption{Tracking a complex signal: (a) Open-loop control; (b) Closed-loop control; (c) Control errors; (d) Feedforward and feedback control outputs. The reference signal is $6.5\sum_{i=1}^{5} \sin\left(2\pi f_i t - \pi/2\right)+40.5$ [deg], where $t \in [0,100]$ is the time [s], with $f_1 = 0.12$, $f_2 = 0.04$, $f_3 = 0.31$, $f_4 = 0.29$, and $f_5 = 0.25$.}
    \label{fig: complex}
    \vspace{-0.2cm}
\end{figure}

Additional experiments were conducted by tracking a chirp signal and a complex signal, with results shown in Fig.~\ref{fig: sweep} and Fig.~\ref{fig: complex}. Results from Fig.~\ref{fig: sweep} show that although the bending-motion tracking error increases with the desired signal's frequency, it remains within $\pm 8$ [deg] in the closed-loop system without detectable phase lag at high-frequency zones. Results in Fig.~\ref{fig: complex} exhibit the FPRC model's capability to control the actuator's bending following a varying signal. The numerical comparison of tracking errors in RMSE is shown in Table~\ref{tab: exp rmse}.

\begin{table}[t]
\centering
\caption{RMSE of tracking different signals with different methods.}
\label{tab: exp rmse}
\renewcommand{\arraystretch}{0.6}
\begin{tabular}{lcccc}
\toprule
{Control system} & {Fig.~\ref{fig: 0.2}} & {Fig.~\ref{fig: 0.5}} & {Fig.~\ref{fig: sweep}} & {Fig.~\ref{fig: complex}} \\
\midrule
FPRC      & 2.04 [deg]  & 2.25 [deg] & 2.67 [deg] & 1.70 [deg] \\
FPRC+PD   & 0.795 [deg] & 1.16 [deg] & 1.70 [deg] & 0.608 [deg]  \\
\bottomrule
\end{tabular}
\vspace{-0.2cm}
\end{table}

\begin{figure}[t]
    \centering
\includegraphics[width=0.8\linewidth]{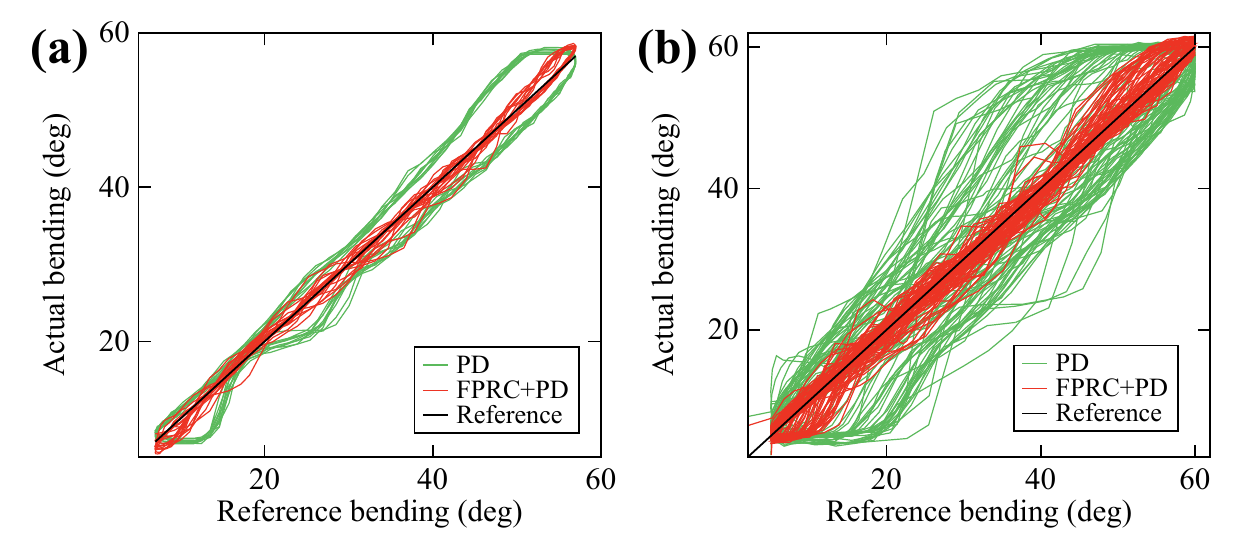}
    \caption{Control results comparison between FPRC+PD and PD: (a) Tracking the reference signal in Fig.~\ref{fig: 0.5}; (b) Tracking the reference signal in Fig.~\ref{fig: sweep}.}
    \label{fig: control_hys_loops}
    \vspace{-0.2cm}
\end{figure}

To demonstrate the FPRC+PD control system's superior performance over the standalone PD controller, we used the PD controller to track the signals shown in Fig.~\ref{fig: 0.5} and Fig.~\ref{fig: sweep} and compared its tracking outcomes with that from the FPRC+PD system. The results are shown in Fig.~\ref{fig: control_hys_loops}, where the PD controller exhibits noticeable hysteresis loops in the reference-to-actual relationship. In contrast, including the FPRC feedforward model largely mitigates such loops.

\begin{figure}[t]
    \centering
\includegraphics[width=0.8\linewidth]{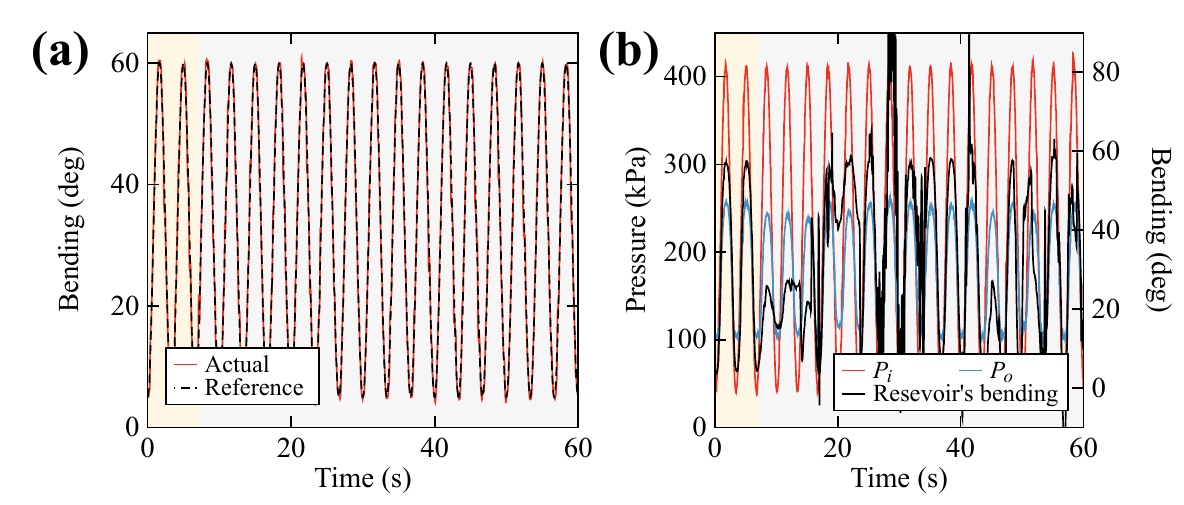}
    \caption{Tracking a 0.3 Hz sine signal with external disturbance: (a) Closed-loop control tracking result of the actuator; (b) Input and output pressures and bending state of the physical reservoir. Results are depicted with background colors of yellow for no disturbance and gray for under manual disturbances.}
    \label{fig: disturbance}
    \vspace{-0.2cm}
\end{figure}

A robustness test was conducted using the FPRC+PD to control the actuator's bending following a 0.3 Hz sine wave. Initially, the pneumatic physical reservoir worked without disturbances for a short period. Subsequently, the physical reservoir was subjected to manually imposed random disturbances. A bending sensor recorded the physical reservoir's bending motion during the experiment to provide an intuitive vision of the disturbance's effects on the physical reservoir. The tracking results and the physical reservoir's bending angle are shown in Fig.~\ref{fig: disturbance} (a) and (b). The experimental results in (b) exhibit that while the physical reservoir's bending was significantly distorted by disturbances, its output $P_o$ remained stable and was not substantially influenced. The tracking results in (a) further indicate that the control system's performance was unaffected by the disturbances applied to the physical reservoir.

The physical reservoir's exceptional environmental robustness can be attributed to the fact that variations in $P_o$ are caused by the compression of the inextensible fabric shield, which is primarily driven by the inflation of the active PAM rather than by the reservoir's actual bending. Consequently, regardless of the reservoir's bending being disturbed, the pressurization $P_i$ of the active PAM compels the fabric to squeeze the passive PAM, thus ensuring the physical reservoir's robust operation.

\section{Conclusion}

This study introduced a PRC-based feedforward hysteresis compensation method, termed the FPRC model, to control a pneumatic soft actuator's bending motion. The proposed method used a dual-PAM pneumatic soft bending actuator as a physical reservoir, with a T-S fuzzy model processing the physical reservoir's readout and generating the feedforward control command. The proposed FPRC model demonstrated comparable model training performance to an ESN model while performing better in the model test phase. The FPRC model required less training time than the ESN model and showed a 97.6\% reduction in the execution time cost during the model test phase than the ESN. Experiments validated the FPRC model's effectiveness in controlling the soft actuator's motion with both open-loop and closed-loop control system applications. The FPRC model's strong robustness against real-world disturbances has also been experimentally verified.

The control systems in this work followed the model-free design, which is easily applicable yet effective. However, the RC's inherent complexity \cite{yan2024emerging, waegeman2012feedback} and the absence of the controlled soft actuator's dynamic model make investigations like the system stability challenging. Future work could explore the use of PRC as a substitute for the ESN in online learning control systems \cite{park2016online, waegeman2012feedback}, where the stability analysis has been discussed. Additionally, as removing the necessary components required by the PRC system (e.g., actuation and feedback units) may compromise its functionality, future work could focus on reducing the system's spatial footprint through miniaturization or integrating actuation and feedback units into a compact architecture. While replacing the pneumatic physical reservoir with more compact and scalable alternatives such as memristor-based PRCs \cite{moon2019temporal} can help alleviate the system's bulkiness, cared attention may be required in matching the PRC's timescale with that of the controlled system--e.g., using a pneumatic soft body PRC for controlling a pneumatic soft actuator--to ensure the PRC's computational performance \cite{yan2024emerging}.

\end{document}